\documentclass[conference]{IEEEtran}

\usepackage{amsmath,amsfonts,amssymb}
\usepackage{authblk}
\usepackage{graphicx}
\usepackage{tabularx}
\usepackage{lipsum}
\usepackage{cite}
\usepackage{array}
\usepackage{geometry}
\usepackage[colorlinks=true, urlcolor=blue, linkcolor=red]{hyperref}
\usepackage{booktabs}
\usepackage{tablefootnote}
\usepackage[small, tableposition=top]{caption} 
\usepackage{amsmath, mathtools, amssymb, algorithm, algpseudocode, caption, subcaption}

\title{Adaptive Meta-Learning for Robust Deepfake Detection: A Multi-Agent Framework to Data Drift and Model Generalization}
\author{Dinesh Srivasthav P, Badri Narayan Subudhi\\
Indian Institute of Technology Jammu\\
Jammu and Kashmir, India\\
 \texttt{\{2022pai9055, subudhi.badri\}@iitjammu.ac.in}}
  
\begin{document}

\maketitle
\begin{abstract}
Pioneering advancements in artificial intelligence, especially in generative AI, have enabled significant possibilities for content creation, but also led to widespread misinformation and false content. The growing sophistication and realism of deepfakes is raising concerns about privacy invasion, identity theft, and has societal, business impacts, including reputational damage and financial loss. Many deepfake detectors have been developed to tackle this problem. Nevertheless, as for every AI model, the deepfake detectors face the wrath of lack of considerable generalization to unseen scenarios and cross-domain deepfakes. Additionally, adversarial robustness is another critical challenge, as detectors drastically underperform with the slightest imperceptible change. Most state-of-the-art detectors are trained on static datasets and lack the ability to adapt to emerging deepfake attack trends. These three crucial challenges though hold paramount importance for reliability in practise, particularly in the deepfakes domain, are also the problems with any other AI application. This paper proposes an adversarial meta-learning algorithm using task-specific adaptive sample synthesis and consistency regularization, in a refinement phase. By focussing on the classifier's strengths and weaknesses, it boosts both robustness and generalization of the model. Additionally, the paper introduces a hierarchical multi-agent retrieval-augmented generation workflow with a sample synthesis module to dynamically adapt the model to new data trends by generating custom deepfake samples. The paper further presents a framework integrating the meta-learning algorithm with the hierarchical multi-agent workflow, offering a holistic solution for enhancing generalization, robustness, and adaptability. Experimental results demonstrate the model's consistent performance across various datasets, outperforming the models in comparison. The associated code is available \href{https://github.com/dineshsrivasthav/adaptive_meta_learning_with_multi_agent_framework}{here}.
\end{abstract}
\section{\textbf{Introduction}}
Deepfakes have become one of the most concerning possibilities of artificial intelligence. They refer to highly realistic falsification of information or media corresponding to one or more modalities such as image, video, audio, and text, with the help of advanced deep learning and generative models. Though media manipulations, especially digital image modifications such as image splicing, colorization, use of filters, image superimposition, and so on were existing from several decades, the intent was usually or mostly for content enhancement. The term ‘deepfake’ was coined in 2017\footnote{https://mitsloan.mit.edu/ideas-made-to-matter/deepfakes-explained} referring to the creation of convincingly realistic fake content post a Reddit user created celebrity face swaps. Despite its malicious beginnings, deepfakes initially, gained attention for their use in entertainment, art, and harmless fun. However, their potential for misuse quickly became apparent, as deepfakes were increasingly used for various malicious purposes, with the rapid evolution of deepfake technology. 
Several varieties of deepfakes are in play, each leveraging different techniques to create convincing fake content across various media formats. Some prominent examples are listed in Table I.

\begin{table*}[ht]
\centering
\begin{tabular}{|>{\raggedright\arraybackslash}p{3cm}|>{\raggedright\arraybackslash}p{2.5cm}|>{\raggedright\arraybackslash}p{2.5cm}|>{\raggedright\arraybackslash}p{2.5cm}|>{\raggedright\arraybackslash}p{3cm}|}
\hline
\textbf{Image Deepfake} & \textbf{Video Deepfake} & \textbf{Audio Deepfake} & \textbf{Text Deepfake} & \textbf{Multimodal Deepfake} \\
\hline
\begin{itemize}
    \item Face Swap
    \item Expression Swap
    \item Attribute Manipulation
    \item Face Synthesis
    \item Face Reenactment
    \item Lip-Syncing
    \item Body Manipulation/swap
    \item Background Manipulation
    \item Background people manipulation
    \item Foreground and background manipulation \& the variants
    \item Style Transfer
    \item Gender manipulation
    \item Age Progression/Regression
\end{itemize} & 
\begin{itemize}
    \item All image deepfake methods applied to videos
    \item Lip-Syncing
    \item Puppet Master
    \item Whole-Head Synthesis (Head Puppetry)
    \item Whole-Body Synthesis (Full body puppetry)
\end{itemize} & 
\begin{itemize}
    \item Text-to-Speech
    \item Voice Conversion
    \item Speech Synthesis
    \item Voice Cloning or Impersonation
    \item Audio Reenactment
    \item Audio Style Transfer
    \item Audio Manipulation
    \item Audio Splicing
    \item Audio Dubbing
    \item Audio Gender Change
\end{itemize} & 
\begin{itemize}
    \item Writing style impersonation
    \item Synthetic text / text-to-text synthesis
    \item Content manipulation/ falsification/ fabrication
\end{itemize} & 
\begin{itemize}
    \item Audio-visual Synthesis (combining audio and video deepfakes and variants)
    \item Text to image synthesis
    \item Text to video synthesis
    \item Text to audio synthesis
\end{itemize} \\
\hline
\end{tabular}
\captionof{table}{Types of Deepfakes}
\end{table*}

More advanced deepfakes combine two or more modalities to create multimodal deepfake content. For instance, a single video could feature a synthesized face, a manipulated background, a cloned voice, and modified lip-syncing, all working together to create an extremely convincing false narrative. The complexity and realism of such multimodal deepfakes make them extremely challenging to detect. This is raising serious concerns about trustworthiness and authentication of digital content, posing significant threat to society due to their ability and high impact in seamlessly spreading misinformation, and invading privacy of people without consent, causing emotional distress with the creation of fake personas tampering individual credibility with theft of identity. The widespread highly convincing fake content known no bounds is also prominently being used to deceive viewers, manipulate public opinion on critical matters including political contexts, and frame individuals and celebrities for incidents or crimes they did not commit. Deepfakes also have a severe impact on businesses such as damaging brand reputation, erosion of trust among customers, monetary loss, loss of market, leading to involvement in fraudulent transactions, and so on.\\
With the growing sophistication of deepfakes, fueled by the proliferation of generative AI, and democratization of a multitude of generative tools, it is crucial to develop effective detection and prevention methods to mitigate these risks. Addressing these challenges necessitates the development of sophisticated and comprehensive solutions that extend beyond traditional detection methods with a multi-faceted approach. However, with the recent increased use of a combination or ensemble of manipulation methods, and the rapid pace at which numerous new variants of deepfakes are emerging, deepfake detectors are increasingly being defeated in identifying deepfakes, making deepfake detection very problematic and inefficient for practical usage.

Followed by this general overview and the broad problem of deepfake detection, there are specific challenges associated with it as follows:
\begin{itemize}
    \item The problem of lack of generalization in current deepfake detectors which often fail in the case of identifying real-time deepfakes and the deepfakes created by another method (cross-domain), despite having the state-of-the-art or a reasonably good performance when trained upon a good quality benchmark dataset.
    \item In most of the cases, minimal changes to model architectures are being proposed which show certain improvement over the prior state-of-the-art model on a specific dataset, but when this better performing model is used elsewhere, its performance gets significantly dropped.
    \item Also, these models are often just trained on the specific bench-marked datasets and the accuracy improvements are reported on the same. However, they are not often adversarially trained or tested due to which they are practically not usable, as they are very sensitive to minute imperceptible changes and are prone to malicious attacks with slightest tweaks.
    \item Keeping up with dynamic data drift with rapid evolution of new deepfakes, and adapting to new tactics is a concerning challenge. The state-of-the-art models are trained on pre-curated static samples and thus, have no way to get updated with the changes in deepfake patterns that are rapidly evolving. Finetuning the model on a new dataset is the usual way followed, however, eventually, this ends up with the same problem. Hence, a dynamic model updation strategy is needed. 
\end{itemize}
In this paper, we attempt to address the above challenges, by proposing an adversarial meta-learning algorithm with multi-agent framework for generalization, robustness, data synthesis, and adaption to dynamic data drift. The main contributions of this paper are as follows:
\begin{itemize}
    \item An adversarial meta-learning algorithm amalgamating the prospects of task-specific adaptive sample synthesis, and consistency regularization, in a few-shot computationally non-intensive setting, with a refinement phase boosting the generalization and robustness of the model.
    \item A formula $M_{adaptive}$ to identify the samples where the model is very confident, and the samples where the model is greatly struggling, using prediction probability, margin, classification sign, and entropy.
    \item A unified loss function using weighted contrastive and margin ranking loss.
    \item A hierarchical multi-agent retrieval-augmented generation (RAG) workflow with a sample synthesis module for collecting and generating custom synthetic deepfake samples for dynamic model training. To the best of our knowledge, this is the first work to implement such a workflow with agents for collecting real-time information, synthesize deepfake attack patterns, and corresponding few-shot prompts, followed by the custom sample generation.
    \item A holistic framework integrating the proposed meta-learning algorithm with the hierarchical multi-agent RAG workflow for an end-to-end deepfake detection architecture enhancing generalization, robustness, and adaptability.
\end{itemize}
The rest of the paper is organized as follows. Section 2 discusses about the related works for addressing the aforementioned challenges, and their limitations. Section 3 details the proposed methodology, and Section 4 describes the implementation and experimentation of the framework along with the results. Finally, Section 5 concludes the proposed work along with discussion regarding the future scope of work. 

\section{\textbf{Related work}}
Over the years, with the rise in sophistication in deepfake generation, deepfake detection has evolved from employing classic machine learning to advanced generative neural network (GAN) and transformer-based approaches. 

Post the early primitive methods based on traditional image processing techniques, classic machine learning algorithms such as support vector machine, and decision trees were trained to differentiate between real and manipulated content based on basic visual cues for images. However, with the growing sophistication of deepfakes, classic machine learning based methods and other primitive techniques were ineffective in detecting deepfakes, due to which there was a quick transition to employing deep learning methods. Convolutional neural networks (CNN) were prominent for a long period in identifying manipulated visual content due to their ability to automatically learn spatial hierarchies of features. These efforts focussed on detecting inconsistencies in facial landmarks, texture patterns, and other low-level image features.
As the complexity of deepfakes increased, deeper CNN architectures such as ResNet, AlexNet, and VGG16 were used. These models were pre-trained on large datasets like ImageNet, and demonstrated enhanced detection capabilities when fine-tuned on deepfake-specific datasets.

As the limitations of CNN-based methods became evident, researchers turned towards GANs both for generating and detecting deepfakes. GAN-based detection models, such as those using CycleGAN, StarGAN, BigGAN, and StyleGAN, focused on detecting spatial inconsistencies and temporal anomalies in images and video sequences. However, GAN-based methods often require large amounts of computational resources, limiting their practical applicability. Despite this challenge, most datasets available today have used some form of GAN, besides many other generative models such as diffusion-based, for data generation, customization, and enhancement due to their generating capabilities. 

Transformer-based models, inspired by their success in natural language processing, were adapted for deepfake detection to capture long-range dependencies in images, videos, audio, text. Vision Transformers (ViTs), for instance, demonstrated the ability to capture global context better than CNNs and other earlier models. There are several works based on variants of ViT and other transformer architectures that have demonstrated state-of-the-art results. While transformers can offer superior performance, they come with challenges related to scalability, computational cost, and the need for extensive training data. Nevertheless, due to their superior performance, transformer and attention-based architectures are still considered the best and being used for most works.

Multimodal approaches have emerged as a promising direction, integrating information from various modalities, such as video, audio, and text, etc., to enhance detection capabilities and adaptation to practical usecases. However, despite these advancements, multimodal methods are still commonly challenged by issues of data format dependency, scalability, appropriate rich data availability, and multimodal processing. For instance, an Audio-Visual model need an audio-visual input for inference and may not work for either modality alone, as most approaches are designed with feature fusion, inter-feature dependency, cross-modal feature cues, and so on. Nevertheless, there are also approaches that are designed to work with one or more modalities alleviating the problem.

Despite the increased sophistication of deepfake detectors, there are still crucial challenges where active research and development is going on and further needed. Some of these key challenges are discussed below:\\
\begin{figure*}[ht]
  \centering
  \includegraphics[width=\linewidth]{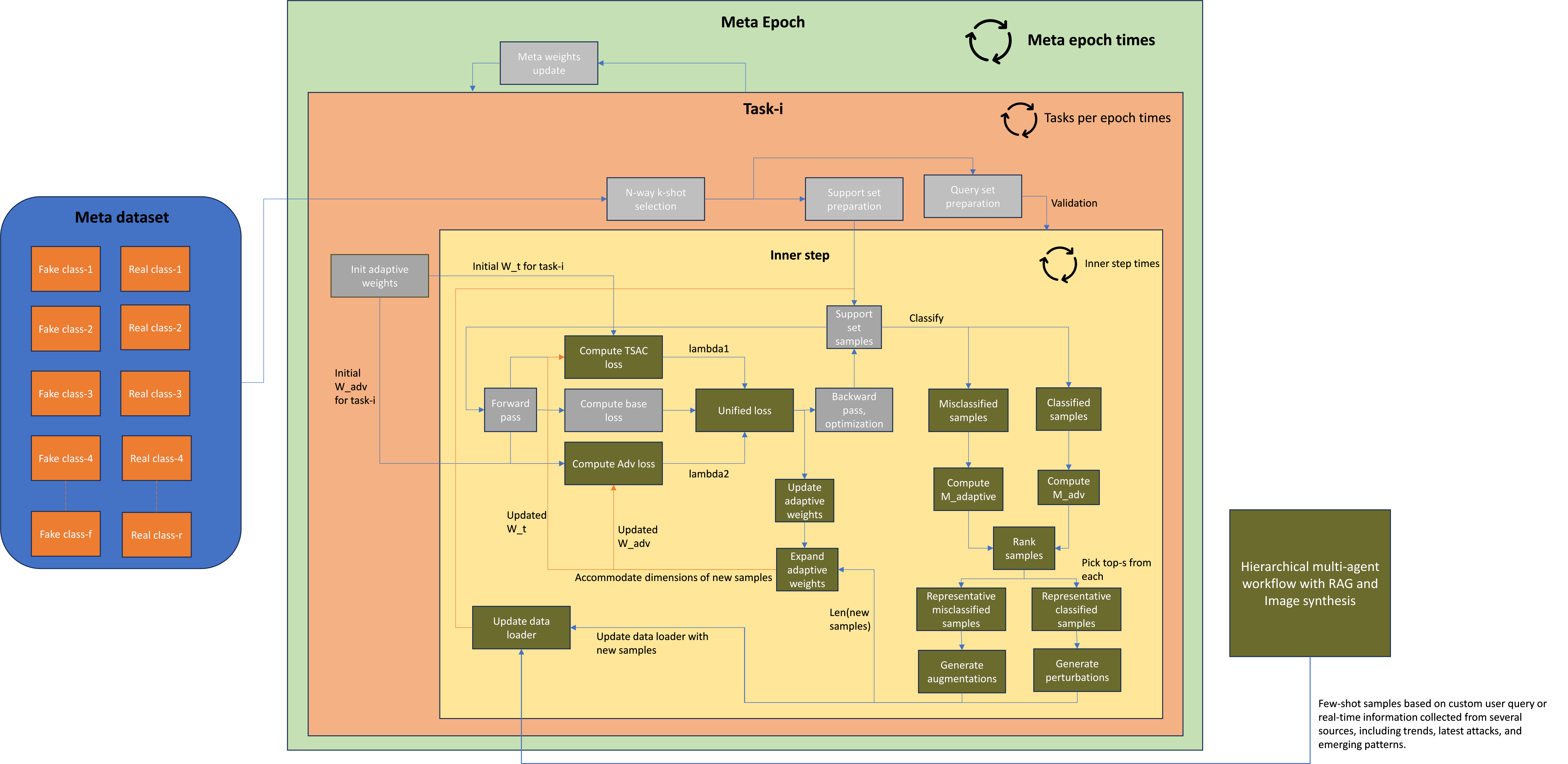}
  \caption{Proposed framework (all the olive green blocks are the additions we added to the Reptile algorithm)}
  \label{fig:framework_1}
\end{figure*}
\textbf{Generalization:}
While generalization is an overarching problem for many if not most of the artificial intelligence, machine learning, deep learning use cases, it holds a matter of critical concern in the deepfake domain, as one cannot anticipate what the target attack pattern would be, and train a model just for that.  
In order to improve generalization and adaptation of the model to a wider variety of samples, many approaches such as transfer learning, fine tuning, knowledge distillation, domain adaptation, and so on are present \cite{ article_tl, 9523015, 10444215}.
The cross-generator deepfake image classification performance reported by Song et al. \cite{ song2023robustnessgeneralizabilitydeepfakedetection} are in the range of 38\% to 59\% where a state-of-the-art RECCE method which was trained on the benchmarked dataset of FaceForensics++ was adopted and validated on the DeepFakeFace dataset in a cross-generator image classification setting. 

Though each of these approaches have their own advantages, and established tailored applications, they may not be very effective for broader generalization despite being quite useful and efficient for small-scale or specific task adaption. This could be apparently because, in each case, a pre-trained model is just adapted to a particular new task using its parametric memory, but ideally, it is not learning to generalize to a broader range of tasks. Repeated fine-tuning or transfer learning for numerous tasks would not be very efficient. Meta learning helps in generalization for a bigger spectrum of tasks, and is often trained that way upon multiple tasks sampled for each meta epoch, due to which the model’s weights will move towards an optimal set of values from which the model can be quickly adapted to many tasks. 

However, the study by Boquan Li et al. \cite{li2024generalizabledeepfakeimagedetectors} confirms that image deepfake detectors do not generalize well in zero-shot settings from their experimentation with six datasets and five state-of-the-art deepfake detectors in various combinations across multiple settings. Besides, they reported that the models were learning unwanted features that were not discriminative enough to generalize well to unseen scenarios.

The survey by Passos et al. \cite{ Passos_2024} done during the period of 2018-2024 reviewing 89 papers emphasized the poor performance of models on cross-dataset tests. Even the well performing models trained on benchmarked datasets reported drastic decrease in performance when validated on different deepfake datasets.

The results from the experimental study by Kamat et al. \cite{ Kamat_2023_ICCV} proves yet again that the deepfake detection methods fail in different scenarios. They experimented with three state-of-the-art detectors trained on varied artifacts when evaluated on four newer deepfake types where one of the evaluated models have shown a drop in 41\% drop in AUC, while the other two have dropped 24\% and 16\%. This not only highlights the problem of generalization even in the state-of-the-art methods, but also sheds light on how rapid are the deepfake attack patterns and synthesis techniques evolving, rendering even the best models ineffective.

\begin{figure*}[ht]
  \centering
  \includegraphics[width=\linewidth]{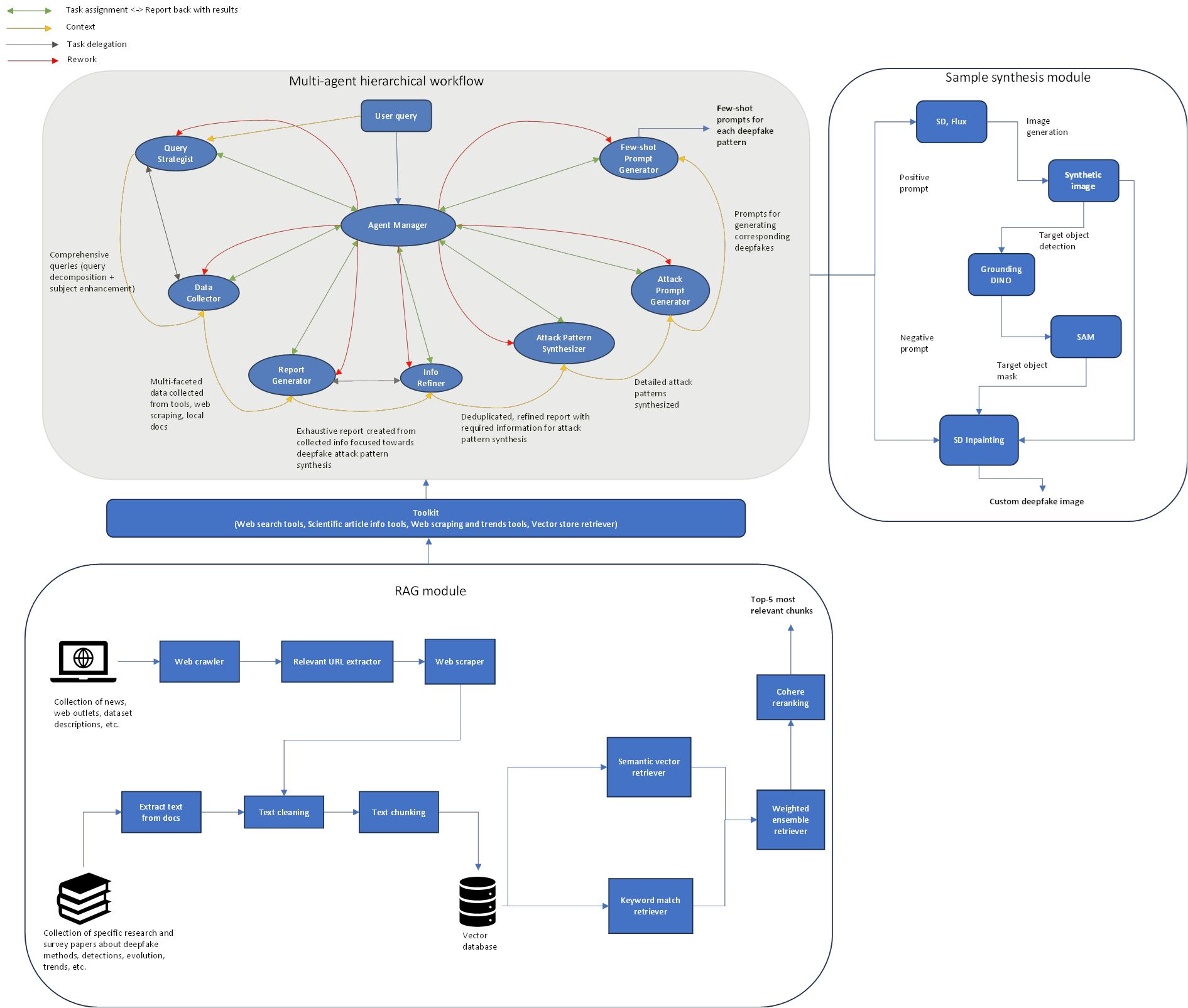}
  \caption{Hierarchical multi-agent workflow for custom deepfake sample synthesis}
  \label{fig:workflow_2}
\end{figure*}

\textbf{Adversarial robustness:}
Marija et al. \cite{10495703} investigated if the image deepfake detectors are susceptible to the black-box attacks created by denoising diffusion models, and noted that even if the deepfake reconstruction process includes a single denoising diffusion step, without showing any notable perceptual changes, the performance and the detection capability of the deepfake detectors have greatly come down. It was further noted that detectors solely trained on the train samples were more vulnerable than when also trained on attack examples which enhanced robustness.
Carlini et al.\cite{ 9150604} proved in their work that even state-of-the-art image deepfake detectors with a consistent 0.95 AUC on images synthesized by several generator methods can be nullified with a white-box attack of seed perturbation where just the lowest bit of each pixel was flipped to bring down the model to nothing. They even proved that this would be the similar results even with a black-box attack where no information of the detector is known – it can still be dropped to quite low, 0.22 AUC as reported. They have also showcased several other attack case studies consistently proving the impact of adversarial perturbations with only minute or almost with no perceptible image alternations. This very much highlights how important adversarial robustness really is.  
Likewise, Hou et al. \cite{hou2023evadingdeepfakedetectorsadversarial} have devised an adversarial attack statistical consistency attack, based on the statistical properties of original and deepfake images. This attack tries to create adversarial samples whose feature distributions are very close to the natural images, thereby failing the detectors. They have established this with multiple scenarios including both white- and black-box attacks with six different detectors on four datasets.
To enhance the adversarial robustness against the adversarial attacks like above, there are many works proposed such as \cite{10022079, hooda2023d4detectionadversarialdiffusion, 10.1007/978-3-031-53311-2_37}. However, there are even more ways \cite{9522903, galdi20242dmalafideadversarialattacksface, 9956543, 9956520, 10035845} to design robust adversarial attacks that are strong enough to fool the best of the best models.

\textbf{Data drift:}
As new generative models and deepfake synthesis techniques evolve, data drift or a shift in the data distribution is bound to happen. Especially, with the rapid pace at which new attack patterns are evolving, data drift is a continuous phenomenon. Approaches like online learning, continual learning, etc., are usually used to make the machine learning models keep up with changing patterns in the incoming data. However, most deepfake detection models and approaches rely on static training sets, and are not widely observed from the literature to update continuously or periodically. Due to this, a severe difference between the training data used for model training and the characteristics and nuances of the incoming new data samples will arise, making the deepfake detectors incompetent. 

Kamat et al. \cite{ Kamat_2023_ICCV} mentions that the existing out-of-the-domain evaluation datasets are similar to the data used for model training and do not cope up well with the advancements of deepfake attack patterns and synthesis techniques, thereby leaving the deepfake detectors inefficient for different scenarios and vulnerable.

Tassone et al. \cite{TASSONE2024104143} highlights that data drift can greatly limit the applicability of deepfake detectors, and proposed an approach based on continual learning with MlOps CI/CD pipeline from a particular set of data sources. However, this approach is built on a few requirements such as the various tasks should come as batches to the system and they should be grouped based on their similarity. Their experimentations revealed that these factors are crucial for the model’s performance and are required to be maintained to get optimal results in their setup. Moreover, the approach suggested manual intervention of forensic experts and journalists to analyse the collected data and prepare the model training data.

\begin{figure*}[ht]
  \centering
  \includegraphics[width=\linewidth]{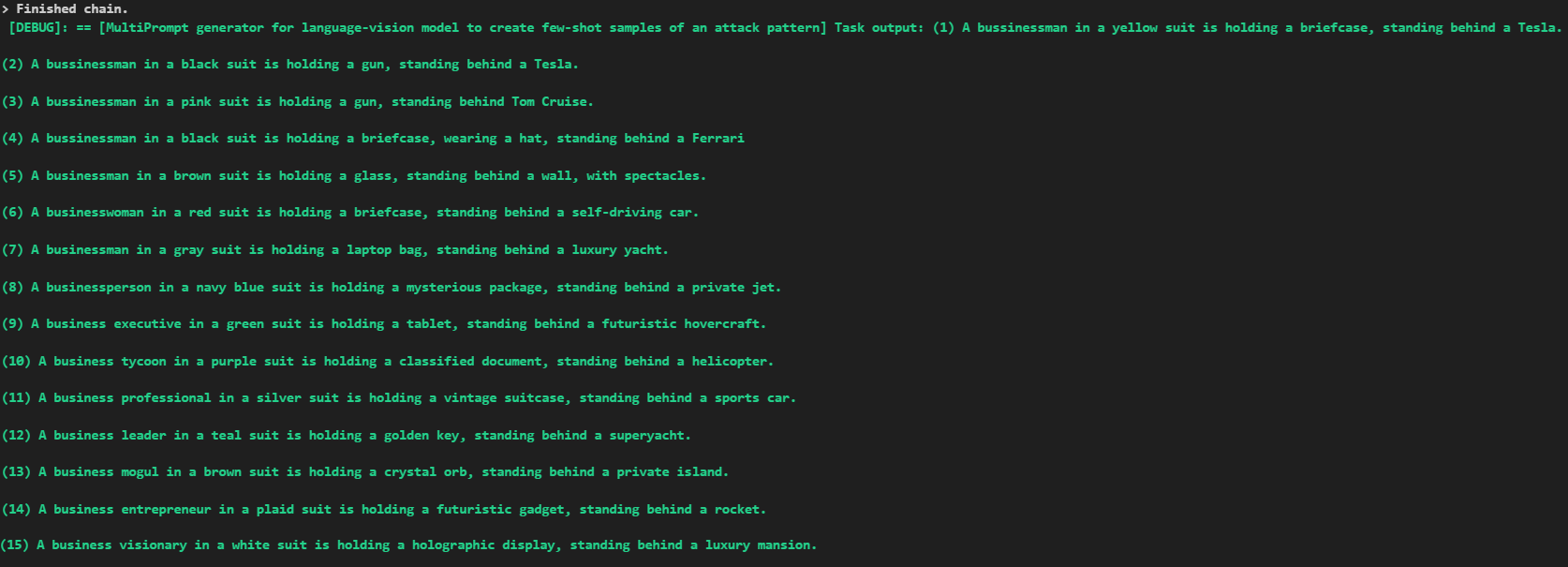}
  \caption{Output of few-shot prompts from the multi-agent hierarchical workflow for the positive prompt: "A bussinessman in a black suit is holding a briefcase, standing behind a Tesla"}
  \label{fig:prompts_3}
\end{figure*}

\section{\textbf{Methodology}}
To address the highlighted challenges faced by the current detectors, the following framework is being proposed. It primarily focuses on enhancing model generalization with adversarial robustness, and coping up with data drift. It has two major modules as depicted in Figure 1: Meta learning module, and Hierarchical multi-agent workflow with RAG and Image synthesis. We propose a novel meta learning algorithm with an added refinement phase over the Reptile algorithm \cite{nichol2018firstordermetalearningalgorithms}, amalgamating the prospects of task-specific adaptive sample synthesis with consistency regularization, where concerned augmented samples are generated to aid the instances where the model is struggling with misclassifications, and task-specific adaptive perturbed samples will be used with training samples to boost the robustness of the model against several combinations of adversaries. The functioning of these modules are elaborated below.

\subsection{\textbf{Hierarchical multi-agent workflow with RAG and Image synthesis}}
This workflow is intended to tackle the problem of dynamic data drift, where the model trained on static data would find it difficult to identify the rapidly changing attack patterns and variants. Therefore, the following workflow is designed to generate deepfake samples based on the custom user query, or based on the real-time information collected from several sources, including trends, latest attacks, and emerging patterns. The workflow has three broad modules: RAG, Multi-agent hierarchical workflow, and Sample synthesis, as depicted in Figure 2, which are described below, and can be triggered periodically to keep up with the emerging trends.\\

\subsubsection{\textbf{RAG module}}
The RAG module helps in retrieving relevant contextual information from various constantly updating sources to provide the corresponding information to the agents present in the multi-agent hierarchical workflow, for them to accordingly synthesize concerned deepfake attack patterns. The knowledge base for the RAG module primarily relies on two sources. The first source is a local collection of specific research papers, survey papers, and technical articles, \cite{Nguyen_2022, Mirsky_2021, jimaging9010018, mittal2022facedetectionmodelsbiased, 10204714, narayan2022deephydeepfakephylogeny, 10440475, wang2023deterdetectingeditedregions, song2023robustnessgeneralizabilitydeepfakedetection, Wang2024LinguisticPO, 9897972, 10208982, Korshunov2018DeepFakesAN, 9156378, 9133490, Li2019CelebDFAL, 9010912, 9578592, 9484387, Zi2020WildDeepfakeAC, 8683164, 9931802, 10469620, inbook, 10285057, 9712265, 10112518, 10448328, 10448251, inproceedings, 10646853, Naitali2023DeepfakeAG} spanning over a few years in the active times of deepfakes. These papers provide holistic information about multifaceted aspects of deepfakes such as the evolution of deepfakes, attack patterns, and variants, generation methods, detection methods, how deepfakes are evading detectors, scientific developments, threat landscapes, emerging trends, and challenges among others. The text from these articles is extracted, processed, chunked, and embedded to the vector database. The second source is a collection of web sources \footnote{https://www.independent.co.uk/topic/deepfake,
https://wionews.com/deep-fake,
https://www.ndtv.com/topic/deepfake-news,
https://thehill.com/social-tags/deepfake/,
https://www.france24.com/en/tag/deepfakes/,
https://economictimes.indiatimes.com/topic/deepfake,
https://www.cbsnews.com/tag/deepfake/2/,
https://www.miragenews.com/tag/deepfake/,
https://paperswithcode.com/datasets?task=deepfake-detection,
https://theconversation.com/us/topics/deepfakes-50460
} primarily of news, web outlets to get the latest happenings in deepfakes, deepfake dataset descriptions to understand the attack patterns of current focus. As these web outlets are not direct articles, but instead host a bunch of articles, news, etc., it is therefore, important to first crawl the web outlet pages, identify the relevant URLs from them, as there can be many irrelevant links, text, and other data present. We used FireCrawlAI\footnote{https://www.firecrawl.dev/} to crawl the web outlets, and used keyword-based extraction to fetch the relevant URLs from them. Post a list of relevant URLs are obtained, a web scraper is deployed to scrape the corresponding URLs and get the relevant information, which is then integrated into the previous pipeline for data cleaning, chunking, and embedding. The data cleaning step also includes a data deduplication, to avoid chunking and storing contextually similar content.\\
When an agent invokes the RAG module with a concerned thought, or query, we employed a hybrid search to retrieve the relevant chunks of information from the vector database. The hybrid search is coordinated by a Weighted ensemble retriever that includes two retrievers: Semantic vector retriever with a weight of 0.7, to retrieve the contextually similar chunks based on vector similarity search. Keyword match retriever, with a weight of 0.3, to fetch the chunks having matching keywords from the query. The retrievals of the Weighted ensemble retriever will be reranked  using Cohere reranking, and the top-5 most relevant chunks will be returned.\\

\begin{figure*}[ht]
  \centering
  \includegraphics[width=\linewidth]{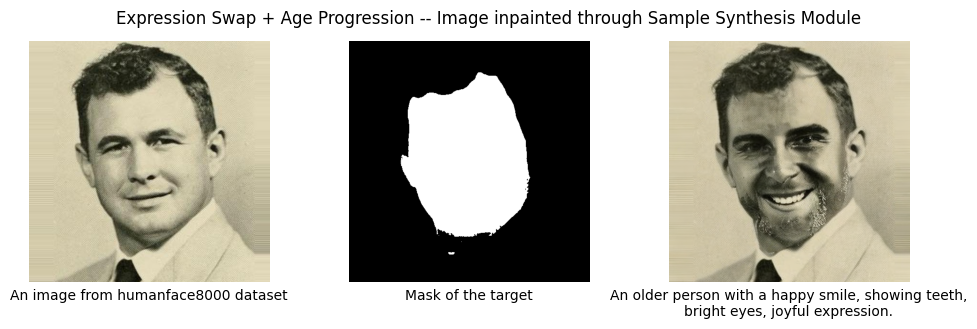}
  \caption{Expression Swap + Age Progression -- An Image inpainted through the Sample Synthesis Module }
  \label{fig:exp_swap_4}
\end{figure*}

\begin{figure*}[ht]
  \centering
  \includegraphics[width=\linewidth]{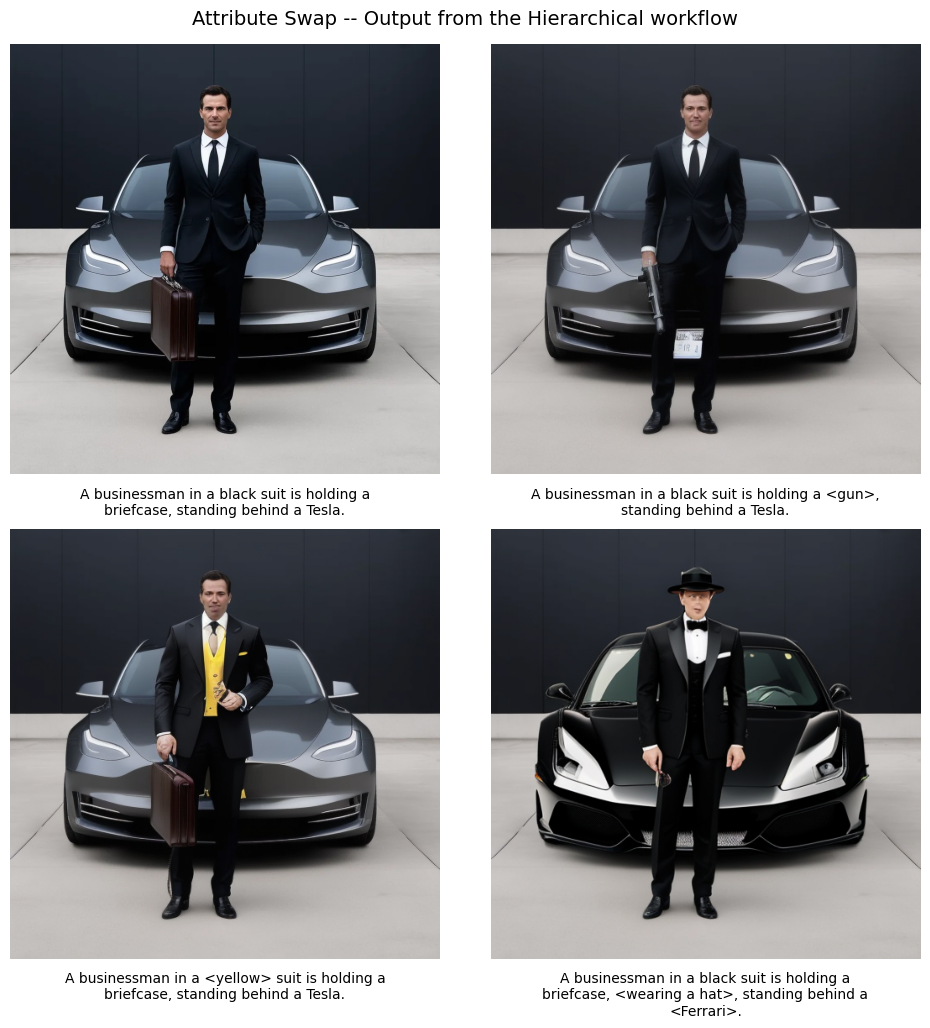}
  \caption{Attribute Swap -- Few-shot samples generated from the Hierarchical workflow module based on few-shot prompts generated by multi-agents as depicted in Figure 3.}
  \label{fig:attr_swap_5}
\end{figure*}

\subsubsection{\textbf{Multi-agent hierarchical workflow}}
The role of this module is to synthesize deepfake attack patterns that might be existing, unforeseen, or hypothetical, and generate language-vision model (LVM) understandable prompts for generating the corresponding image deepfakes. The workflow involves a crew of 7 worker agents and an agent manager, where each agent has a designated role to play, and the agent manager designates the corresponding tasks to the corresponding agents, based on the user query. It supervises the workflow, and the outcomes of each agent, and accordingly asks them to either rework, or delegates it to the next agent with appropriate context. The following are the primary tasks of the agents:
\begin{itemize}
    \item \textbf{Query Strategist:} Given a user query, or an intermediate query generated from the workflow's chain of thought, the Query strategist will create a list of related queries for the given query, like branches or sub-questions, aiding in wholesome coverage of the subject from different perspectives, and sources. The agent, based on the query,  will either decompose it into sub-queries if it is too big or complex, and on the other hand, if it is too simple, or insufficient to trigger further workflow, it will generate comprehensive queries around the subject of seek. It enhances the subject coverage, and helps streamline the queries for better reasoning and associated information collection.
    \item \textbf{Data Collector:} Given a bunch of queries, this agent using its assigned tools collects all the relevant and useful information that eventually helps in synthesizing image deepfake attacks for each of the queries provided. It is also allowed to instill any relevant information from its own pre-trained (parametric) knowledge. Thereby, it curates a significant amount of unstructured information from various sources that becomes the context to the succeeding agent.
    \item \textbf{Report Generator and Info Refiner:} Given the large corpus of unstructured data curated, it is important that it is filtered to remove redundancy of contextually duplicate information. At times, certain amount of not so relevant information might also be collected, which might be useful till or for a particular step, but may not hold much value for further processing of image deepfake attack pattern synthesis, which therefore, has to be discarded. Besides, it is also important to format it into a structured form, as the data is collected from several sources adhering different data format standards. A streamlined flow of relevant information is needed for passing on to the succeeding agents for their tasks. These are the tasks of the Report Generator and Info Refiner agents where the former prepares an exhaustive report of relevant information, and the latter further deduplicates, enhances, and refines it concentrated towards the objective. 
    \item \textbf{Attack Pattern Synthesizer:} Using all the refined information, and its parametric knowledge, this agent synthesizes image deepfake attack patterns which either might be existing, or could be foreseen from the trends and other information collected, or could be completely hypothetical, which still is good enough specifically for this case, unlike other large language model's usecases, as if the model is also trained or made aware of non-existing attack patterns, it would only make the model more robust and prepared.  
    \item \textbf{Attack Prompt Generator:} For synthesizing corresponding deepfake samples using GenAI models or LVMs, precise prompts will be needed rather than lengthy attack patterns. This agent focuses on devising clear, concise and effective prompts to pass on to generative models for sample synthesis, from the detailed attack pattern descriptions. For a given attack pattern, it creates a positive and a negative prompt. One is a prompt to generate a scene with human subject(s) or a portrait of human face. Second is the prompt that describes how the image should look like if it is subjected to the comprehended deepfake attack pattern in a realistic way. For instance, if the first prompt is: 'A person with a happy smile, showing teeth, bright eyes, joyful expression.', the second prompt could be something like: 'A person with a serious face, angry eyes, and a frown, with a sad expression. Here, both the prompts are generating similar images, but with an expression change denoting expression swap deepfake pattern.
    \item \textbf{Few-shot Prompt Generator:} Given the positive and negative prompts for an attack pattern generated by the Prompt Generator agent, this agent is tasked to generate multiple variants of the given prompts to create similar scenarios for few-shot samples of an attack pattern. All these variants follow the same type of deepfake attack pattern (Eg., Expression swap). But they help in creating different images, various scenarios, etc., for creating few-shot examples of the attack pattern. For instance, if the positive prompt is: 'A person with a happy smile, showing teeth, bright eyes, joyful expression.', and the negative prompt is: 'A person with a serious face, angry eyes, and a frown, with a sad expression.', which are generated by the Prompt Generator agent, a sample of the few-shot prompt variants could be: (1) A person with a sad expression, deep eyes, tears rolling. (2) A person with a gentle simple smile, mouth shut. (3) A person with a neutral expression wearing spectacles. (4) A person with raised eyebrows and a surprised look. (5) A person with big open mouth showing a sudden shock. Figure 3 presents an output of few-shot prompts generated from the multi-agent hierarchical workflow for attribute-swap attack pattern.
\end{itemize}

All the agents have access to a toolkit that includes a range of tools useful for fetching a wide range of information from several sources and perspectives, such as web search tools, web scraping tools, relevant scientific article fetching tools, tool to fetch related developments from Google trends, the vector store retriever of the RAG module, among others. The agents use appropriate tools at each step to get relevant information useful to carryout their tasks.\\

\begin{figure}[ht]
  \centering
  \includegraphics[width=\linewidth]{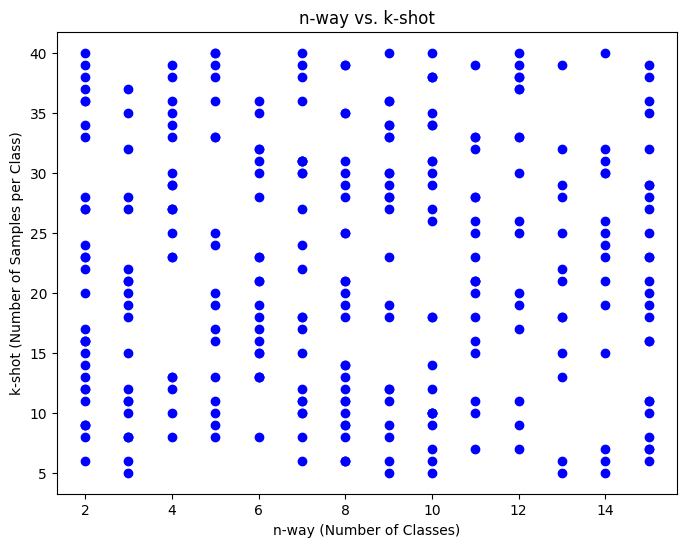}
  \caption{N-Way vs K-Shot}
  \label{fig:img_8}
\end{figure}

\subsubsection{\textbf{Sample synthesis}}
This module aims at synthesizing deepfake images based on the few-shot prompts provided by the multi-agent hierarchical workflow module. The positive prompts are passed to a text-to-image generation models such as Stable Diffusion, Flux, etc. The deepfake manipulation will be applied on this generated synthetic image. The target object from the image is detected using Grounding DINO, and is later segmented using SAM to get the mask of the target. The negative prompt, subject image, and the target mask are passed to the Stable Diffusion Inpainting pipeline to get the deepfake image based on the comprehended deepfake attack pattern. Figure 4 presents an image from the humanface8000 dataset\footnote{https://www.kaggle.com/datasets/bharatadhikari/humanface8000}, that is inpainted with an ensemble of expression swap and age progression deepfake patterns, targeting the face of the person, using this module. Figure 5 represents the few-shot samples generated using the Sample synthesis module, where the few-shot prompts for the same are generated from the multi-agent hierarchical workflow, as depicted in Figure 3. The first sample or image in Figure 5 is the synthetic image generated based on the positive prompt: 'A bussinessman
in a black suit is holding a briefcase, standing behind a Tesla', and the other images here show the attribute manipulations of the synthetic image where one or more attributes have been manipulated according to the synthesized negative few-shot prompts such as the briefcase is swapped with a gun in the second image, the color of suit/vest is changed to yellow from black in the third image, and the Tesla car is replaced with a Ferrari, along with the addition of a hat to the person in the fourth image.\\ 

\begin{figure}[ht]
  \centering
  \includegraphics[width=\linewidth]{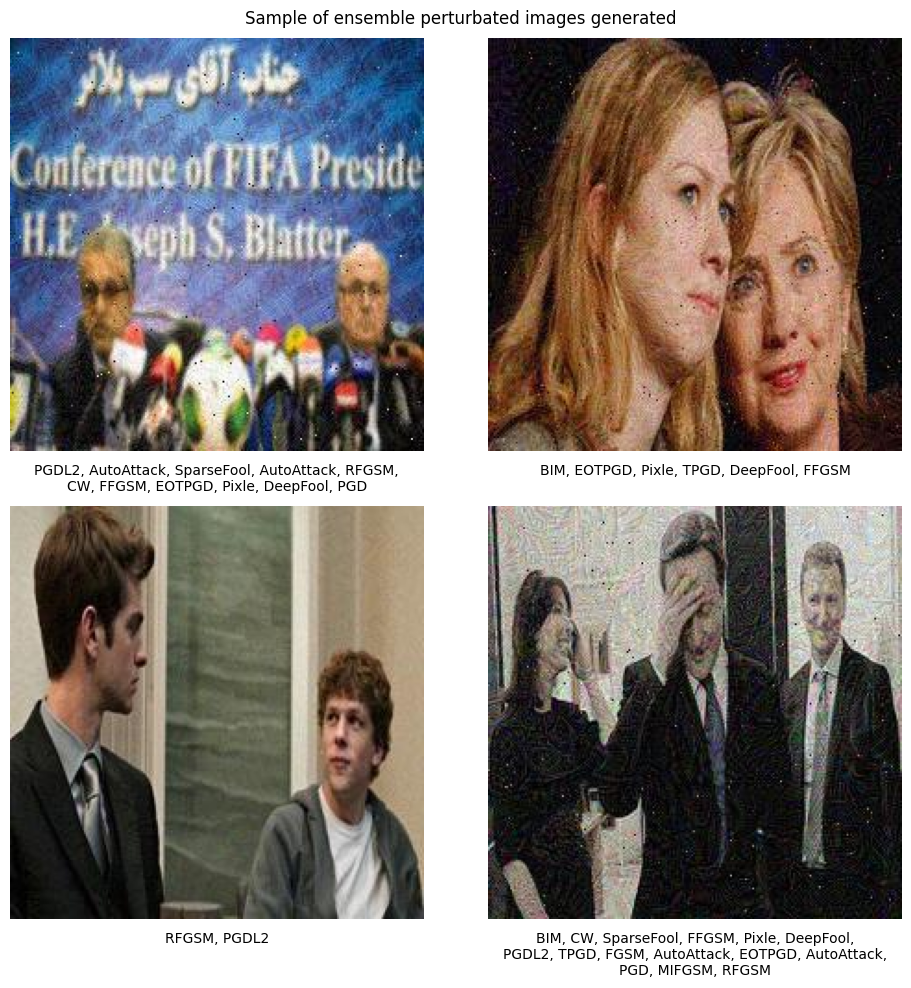}
  \caption{Sample of images post the application of ensemble of perturbations}
  \label{fig:pert_6}
\end{figure}

\subsection{\textbf{Meta-learning algorithm}}
The proposed meta-learning algorithm as represented in Figure 1 concentrates on two directions: 
\begin{itemize}
 \item Identify the samples where the model is facing extreme difficulty in classifying, and create corresponding sample augmentations (one or more as a sequential combination or as an ensemble) which will accordingly be added to support set samples in each inner step, thereby enhancing the model’s sense of decision, thereby, enhancing its generalization capability.
 \item Identify the samples where the model has made extremely confident prediction, and generate corresponding adversarial perturbations (one or more as a sequential combination or as an ensemble) which will accordingly be added to support set samples in each inner step, thereby enhancing the robustness of the model.
\end{itemize}

Therefore, the proposed algorithm focuses on both the strengths and weaknesses of the classifier and trains it accordingly to boost the robustness and generalization.

The intuition behind using task-specific adaptive sample synthesis with consistency regularization is that, in sample synthesis, we generate samples to help the model overcome its difficult challenges. For this, we identify the weaknesses of the model, and generate corresponding samples that are either similar/augmented or are synthetic variants of the weakness scenarios, which supports the model in picking up the deeper intricacies and improves its detection and generalization capabilities.  However, as the very task at hand is deepfake detection, we need to ensure that the model preserves its consistency in detecting real and fake or synthetic images irrespective of whether they are synthetic variants of the original image created to aid overcome its weaknesses, or are actually fake class images. To maintain this, we add an objective term of consistency regularization as weighted contrastive loss. The algorithm is detailed as below.\\

\textbf{Proposed Meta-learning algorithm:\\
Initialization:} Start with a pretrained model.\\
\textbf{Meta-training phase:}\\
Consider there are $E_{meta}$ number of meta epochs (outer loop), and let there be tasks $T = {T_1, T_2,...,T_t}$ for an epoch, where $len(T)$, number of classes in a task $Task-i$, and number of samples per class for task $Task-i$ may vary. Each of these tasks are drawn from two or more classes of the meta dataset $D$. Let each task be trained for $inner\_step$ (inner loop) number of times.\\
\textbf{Refinement phase and unified training:}\\
This is an additional phase we embedded to the training process with the Reptile algorithm as the base, to enhance the generalization and robustness of the model to a variety of unseen samples by combining the prospects of task-specific adaptive sample synthesis, adversary generation, and consistency regularization with meta-learning. 
For meta epochs $E_{meta}$, for every task,
\begin{itemize}
    \item Evaluate the model on the training set of each task and identify the classified and misclassified samples.
    \item Generate synthetic samples for the misclassified samples to improve their generalization, and generate adversarial samples for the correctly classified samples to test the robustness of the model against adversarial perturbations. For generating the respective samples, pick representational samples from the most challenging samples, and most confidently predicted samples respectively identified in each task as follows:
        \begin{itemize}
            \item Compute $M_{adaptive}$ as follows:\\
            $M_{adaptive} = -(p_y - H(p) + Margin - 2*i_{misclassified}*(1+Margin))$\\ where $p_y$ is the prediction probability of true class of sample $y$, $H(p)$ is the entropy of the predicted probabilities, $i_{misclassified}$ is 0, 1 respectively if the corresponding sample is misclassified or not, and $Margin$ is given as $Margin = |(p_{y} - p_{s})|$ where $p_s$ is the highest prediction probability among incorrect classes.  
            \item All the misclassified samples will be ranked in ascending order of $M_{adaptive}$. This will rank the misclassified samples in the order: Wrong predictions with large margin $>$ Wrong predictions with small margin $>$ Right predictions with small margin $>$ Right predictions with large margin. The top-k or a random sample of $n$ from the top-k will be chosen as representational samples for task-specific adaptive sample synthesis. Thus, identifying the most difficult samples for the model to augment.
            \item Representational samples for adversarial synthesis will be chosen in the same manner with the following differences. $M_{adv}$ is calculated as $-M_{adaptive}$ for rightly classified samples instead for misclassified samples. The samples will be ranked in descending order of $M_{adv}$ and the choice is made similarly.
        \end{itemize}
    \item The newly generated samples (synthetic and adversarial) are added to their respective tasks based on the samples using which they were generated, and the model is then trained for this epoch on this updated training task samples.  
    \item For each task, compute the task-specific adaptive consistency loss $L_{TSAC}$ given by 
    \begin{equation}
L_{TSAC} = \sum_{i=1}^{N} W_t^{(i)} \cdot \text{ContrastiveLoss}(f(x_i), f(\hat{x}_i), y_i)
\end{equation} where $W_t^{(i)}$ is the weight associated with task $t$ for the sample $i$, and the $ContrastiveLoss$ is given as\\

$\text{ContrastiveLoss}(f(x_i), f(\hat{x}_i), y_i) = y_i \cdot \left\| f(x_i) - f(\hat{x}_i) \right\|^2 + (1 - y_i) \cdot \max(0, m - \left\| f(x_i) - f(\hat{x}_i) \right\|)^2$\\

where $||f(x_{i}) - f(\hat{x}_{i})||^2$ gives the Euclidean distance between the feature representations of the original and synthetic samples  $f(x_{i}), f(\hat{x}_{i})$ respectively, $y_{i}$ determines if the samples are similar (1) or dissimilar (0), $N$ is the number of task samples, and $m$ is the margin that defines how far apart the dissimilar samples needs to be. 

\item For each task, compute the adversarial loss adapted from Margin ranking loss, given by
\begin{equation}
    \begin{multlined}
    L_{ADV} = \\
    \shoveleft[-0.5cm]{\sum_{i=1}^{N} W_{adv}^{(i)} \cdot \text{MarginRankingLoss}(f(x_i), f(x_i^{adv}), m)}
    \end{multlined}
\end{equation} 
where $W_{adv}^{(i)}$ is the  adversarial task weight for sample $i$, and the $MarginRankingLoss$ is given as\\ 

$\text{MarginRankingLoss}(f(x_i), f(x_i^{adv}), m) = \max(0, m - (f(x_i) - f(x_i^{adv})))$\\

where feature representations of the original and adversarial samples  $f(x_{i}), f(x_i^{adv})$ respectively, $N$ is the number of task samples, and $m$ is the margin parameter. 

\item Aggregate the task-level losses with the base meta loss to form the total loss function as follows.

\begin{equation}
L_{unified} = L_{base} + \lambda_1 L_{TSAC} + \lambda_2L_{ADV}
\end{equation}
where $L_{base}$ is the inner loss of Reptile, and ${\lambda_1, \lambda_2}$ are the hyperparameters.
\item Adaptive weights ${W_t, W_{adv}}$ should be updated within each iteration of the unified training based on the gradient of the unified loss as below.
\begin{equation}
W_{t_{new}} = W_{t_{old}} - \eta \nabla W_{t} L_{unified}(W_{t_{old}})
\end{equation} and
\begin{equation}
W_{adv_{new}} = W_{adv_{old}} - \eta \nabla W_{adv} L_{unified}(W_{adv_{old}})    
\end{equation} respectively, where $\eta$ is the learning rate.    
\end{itemize}
After the refinement, the model is adapted to tasks beyond the primary dataset through few-shot adaptation, and  the performance on the test set of new tasks is evaluated.

\begin{figure}[ht]
  \centering
  \includegraphics[width=\linewidth]{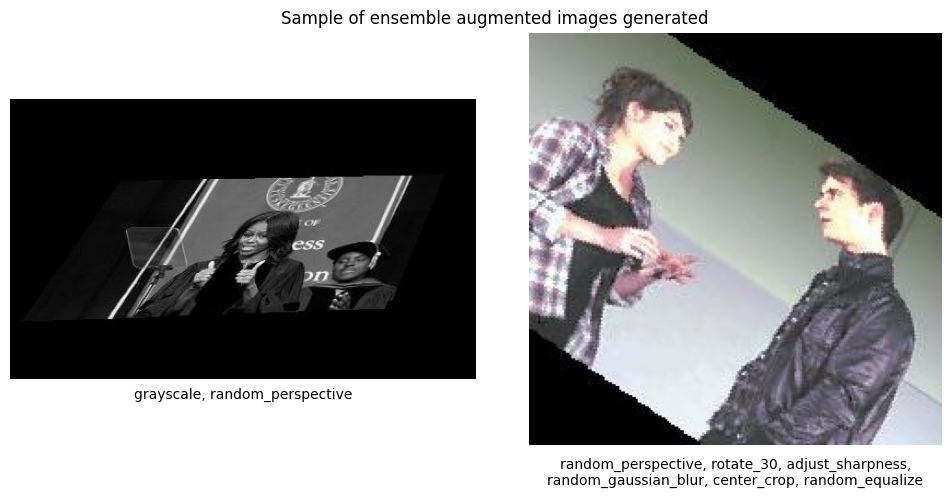}
  \caption{Sample of images post the application of ensemble of augmentations}
  \label{fig:aug_7}
\end{figure}

\section{\textbf{Experimentation \& Results}}

\begin{table*}[h!]
    \centering
    \caption{Meta-Dataset Information}
    \resizebox{\textwidth}{!}{  
    \begin{tabular}{@{}p{2.2cm}p{1cm}p{1.5cm}p{2cm}p{5cm}p{6cm}p{5cm}@{}} 
        \toprule
        \textbf{Dataset} & \textbf{Total Real} & \textbf{Total Fake} & \textbf{Class Count} & \textbf{Deepfake Type} & \textbf{Fake Samples Generation Method} & \textbf{Real Samples Collection Method} \\ 
        \midrule
         DeepFakeFace  \cite{song2023robustnessgeneralizabilitydeepfakedetection}    & 30k      & 90k      & \begin{tabular}[c]{@{}l@{}}Real - 1\\ Fake - 3\\ Total - 4\end{tabular}   & Synthetic images of celebrities   & \begin{tabular}[c]{@{}l@{}}• Stable Diffusion v1.5 (30k)\\ • InsightFace toolbox (30k)\\ • Stable Diffusion Inpainting (30k)\end{tabular} & IMDB-Wiki dataset (30k) \\ 
        \midrule
         DGM    \cite{shao2023dgm4}           & 81.5k    & 125.8k   & \begin{tabular}[c]{@{}l@{}}Real - 4\\ Fake - 4\\ Total - 8\end{tabular}   & Face swap manipulations, face attribute manipulations   & \begin{tabular}[c]{@{}l@{}}• HFGI (28.5k)\\ • Infoswap (44k)\\ • Simswap (22.5k)\\ • StyleClip (30.8k)\end{tabular}   & \begin{tabular}[c]{@{}l@{}}• BBC (15k)\\ • The Guardian (43k)\\ • USA Today (13.9k)\\ • The Washington Post (9.6k)\end{tabular} \\
         \midrule
         iFakeFaceDB \cite{1911.05351}      & -        & 87k (63k+24k)  & \begin{tabular}[c]{@{}l@{}}Fake - 2\\ Total - 2\end{tabular}   & Removing GAN fingerprints from synthetic images   & \begin{tabular}[c]{@{}l@{}}Generated by StyleGAN, and \\transformed with GANprintR,\\ applied on following DBs:\\ • 100k-faces db (24k)\\ • TPDNE db (62k)\end{tabular}   & - \\ 
         \midrule
         CocoGlide\tablefootnote{https://github.com/grip-unina/TruFor} \cite{Guillaro_2023_CVPR} & 513 & 513 & \begin{tabular}[c]{@{}l@{}}Real - 1\\ Fake - 1\\ Total - 2\end{tabular} & Image splicing; Diffusion based & GLIDE (text-guided diffusion model) & COCO dataset\\
         \midrule
         DF40 \cite{DeepfakeBench_YAN_NEURIPS2023} & 4k & 326.5k & \begin{tabular}[c]{@{}l@{}}Real - 3\\ Fake - 14\\ Total - 17\end{tabular}   & Face swapping, Full face synthesis, Face edit & \begin{tabular}[c]{@{}l@{}}• MidJourney (1k)\\ • WhichFaceIsReal (1k)\\ • VQGAN (31.8k)\\ • StyleGAN2 (31.8k)\\ • StyleGAN3 (31.8k)\\ • StyleGANXL (31.8k)\\ • StyleCLIP (31k)\\ • StarGanV2 (2k)\\ • SiT (31.8k)\\ • DiT (31.5k)\\ • RDDM (17.6k)\\ • DDIM (31.8k)\\ • • e4e (50.6k)\\ • CollabDiff  (1k)\end{tabular} & \begin{tabular}[c]{@{}l@{}}• StarGANv2 (2k)\\ • WhichFaceIsReal (1k)\\ • StyleCLIP (1k)\end{tabular}\\
        \midrule
        \midrule
         OpenForensics-based\tablefootnote{https://www.kaggle.com/datasets/manjilkarki/deepfake-and-real-images/data} \cite{ltnghia-ICCV2021}  & 95k & 95k & \begin{tabular}[c]{@{}l@{}}Real - 1\\ Fake - 1\\ Total - 2\end{tabular} & Multi-face forgery; Forged face image synthesis; Face swapping; Various Perturbations and augmentations & GAN; Image transformation; Poisson blending; Color adaptation & Google Open Images\\
        \bottomrule
    \end{tabular}
    }  
\end{table*}

For realising the objective of model generalization, robustness, through meta-learning, we need a very diverse combination of samples, covering a broad spectrum of deepfake types, variants, scenarios, objects, varying complexities, and so on to begin with. This would help in creating heterogeneous tasks for meta-training that would aid in better model generalization.

\begin{figure}[ht]
  \centering
  \includegraphics[width=\linewidth]{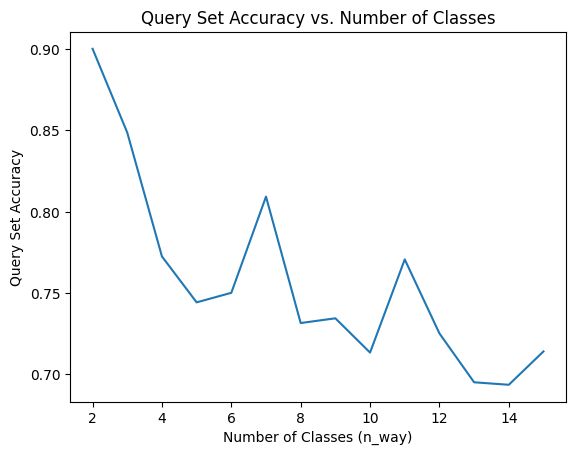}
  \caption{N-Way vs Query Set Accuracy}
  \label{fig:img_9}
\end{figure}

A total of six different datasets were considered as described in Table II. The OpenForensics-based dataset was kept for unseen test evaluation, and the other five datasets were used for meta-training. Each of the individual datasets have two or more classes, and are diverse in nature with respect to the way they are curated, the deepfake type(s) they constitute, the process and generation models through which the fake samples are generated, and the real samples are collected or processed, task and sample diversification, among others. This type of a diverse combination of datasets creates a real-world setting for training, enhancing, and evaluating the generalization of the model to practical scenarios. Together, the meta-dataset constituting of the aforementioned combination of diverse datasets with several classes, totals to 33 classes, of which 9 are real classes, and 24 are fake classes. About 596k samples are part of the train set, 74k samples are part of the validation set, and another 74k samples in the test set.

\begin{figure}[ht]
  \centering
  \includegraphics[width=\linewidth]{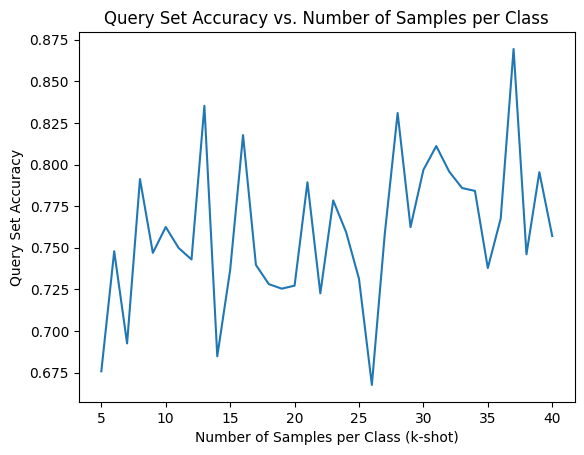}
  \caption{K-Shot vs Query Set Accuracy}
  \label{fig:img_10}
\end{figure}

Though such a mammoth number of varied samples would lead to exceptional training, it requires great hardware and compute power to accomplish it. Specifically, training in a meta-learning setting where each epoch involves the creation of numerous tasks, their respective support and query sets, training the model on each of those tasks, gradient updates, and so on, will require even more computational bandwidth than training a classifier in the typical way. Moreover, meta-training in accordance to the proposed algorithm that further involves a refinement phase with the prospects of task-specific adaptive sample synthesis, consistency regularization, with adversarial robustness will further need greater bunch of resources. Therefore, we have setup our model training in a dynamic few-shot setting, where the number of classes (n-way) and the number of samples per class (k-shot) are indiscriminately drawn from the respective specified intervals in a manner that ensured even distribution. This is depicted in Figure 6. The model architecture chosen is the Co-Scale Conv-Attentional Image Transformer (CoaT) -- 'coat\_lite\_tiny', that was pretrained on the ImageNet-1k. This model was trained with the hyperparameters ${\lambda_1, \lambda_2}$ as 0.5 each, on the meta-dataset for 30 meta epochs (outer loop) where each meta epoch has 10 tasks, and every task is trained for 5 inner steps, thereby counting to 1500 overall training steps on 300 tasks with a heterogeneous combination of classes. At each training step, based on the identified representational samples for the classified and misclassified sample sets computed and ranked based on $M\_{adv}$ and $M\_{adaptive}$, various (sequential combination of one or more as an ensemble) adversarial attacks and sample augmentations as listed in Table-III were respectively applied. Samples of images post the application of ensemble perturbations and augmentations are represented in Figures 7, 8.  These synthesized samples were added to the train sets of the respective classes as described in the proposed algorithm.  The meta-training statistics on the support and query sets are presented in Table IV.
\begin{table}[ht]
    \centering
    \caption{Adversarial attacks and Augmentation methods applied}
    \begin{tabular}{@{}ll@{}}
        \toprule
        \textbf{Adversarial Attacks} & \textbf{Augmentation Methods} \\ 
        \midrule
        FGSM                          & Horizontal Flip               \\ 
        PGD                           & Vertical Flip                 \\ 
        BIM                           & Rotate 30                    \\ 
        CW                            & Rotate 90                    \\ 
        RFGSM                         & Color Jitter                 \\ 
        EOTPGD                        & Grayscale                    \\ 
        TPGD                          & Resize to 128                   \\ 
        FFGSM                         & Center Crop                  \\ 
        MIFGSM                        & Random Equalize              \\ 
        PGDL2                         & Random Solarize              \\ 
        DeepFool                      & Random Perspective           \\ 
        AutoAttack (Linf)            & AugMix                      \\ 
        AutoAttack (L2)              & Auto Contrast                \\ 
        SparseFool                   & Adjust Sharpness             \\ 
        Pixle                         & Random Invert                \\ 
                                    & Elastic Transform             \\ 
                                    & Random Gaussian Blur         \\ 
                                    & Random Resized Crop          \\ 
                                    & Random Erasing             \\ 
        \bottomrule
    \end{tabular}
\end{table}

It is observed that an increase in the number of classes per task is detrimental to query set accuracy, while increase in the number of samples per class in a task has an overall positive effect on the query set accuracy. This is represented in Figures – 9, 10 respectively.
\begin{table}[ht]
    \centering
    \caption{Meta-Training Statistics}
    \begin{tabular}{@{}lc@{}}
        \toprule
        \textbf{Metric}            & \textbf{Value} \\ \midrule
        Support-Set Acc            & 0.7641         \\
        Support-Set Loss           & 1.632          \\
        Query-Set Acc             & 0.7133         \\
        Query-Set Loss             & 1.9817         \\ 
        \bottomrule
    \end{tabular}
\end{table}

Apart from the Meta model, we have trained few other models that are similarly transformer-based, whose architectures are presented in Table-V. These were pretrained on the ImageNet-1k and we then trained them on the DeepFakeFace dataset in the standard way.
\begin{table}[ht]
    \centering
    \caption{Model Architectures}
    \begin{tabular}{@{}ll@{}}
        \toprule
        \textbf{Model}       & \textbf{Architecture}                       \\ 
        \midrule
        Meta model           & coat\_lite\_tiny                           \\ 
        ViT,  ViT\_tl (transfer learning)                  & vit\_tiny\_patch16\_224                    \\ 
        XCiT,  XCiT\_tl (transfer learning)                 & xcit\_nano\_12\_p16\_224                   \\ 
        DeiT                 & deit\_tiny\_distilled\_patch16\_224        \\ 
        Eva2                 & eva02\_tiny\_patch14\_224                   \\ 
        CoaT                 & coat\_lite\_tiny                           \\ 
        Swin\_tl (transfer learning)            & swin\_s3\_tiny\_224                        \\ 
        CNN                  & A 9-layered custom CNN architecture         \\ 
        \bottomrule
    \end{tabular}
\end{table}
To evaluate and compare the Meta model that was few-shot meta trained, with the other models that were fully trained on the DeepFaceFace dataset, we have the following two scenarios:
\begin{table}[ht]
    \centering
    \caption{Performance Metrics on Unseen Dataset -- OpenForensics-based}
    \begin{tabular}{@{}lccc@{}}
        \toprule
        \textbf{Model}  & \textbf{Acc} & \textbf{AUC} & \textbf{F1} \\ \midrule
        Meta model      & 0.6151       & 0.6042       & 0.6319      \\
        ViT             & 0.4714       & 0.5197       & 0.5183      \\
        XCiT            & 0.4495       & 0.4621       & 0.4992      \\
        DeiT            & 0.4525       & 0.4449       & 0.4570      \\
        Eva2            & 0.4780       & 0.4226       & 0.4633      \\
        CoaT            & 0.4649       & 0.4675       & 0.5012      \\ 
        \bottomrule
    \end{tabular}
\end{table}
\begin{itemize}
    \item  Test on a completely unseen dataset (OpenForensics-based), where both the Meta model and the other models have never seen this unseen dataset. The comparison was made using three metrics: Accuracy, AUC, and F1 scores. The results are presented in Table-VI. It is observed that the Meta model has outperformed other fully trained models on all the three metrics. All the other models when tested upon a completely unseen test set, have struggled to reach a random guess performance of 50\% accuracy. In Figure 11, we have presented how the performance of the other models has drastically dropped from over 90\% to about 50\% or less, where the former was when tested on the test set of DeepFakeFace, and the latter was when tested on the unseen OpenForensics-based test set. Additionally, despite the CoaT model, and the Meta model having the same architecture, the CoaT model has got a test accuracy of 46.49\% while the Meta model has got 61.51\% test accuracy on the unseen test set, showing an improvement of about 15\%. However, it is surprising to note that the Swin transformer that was just transfer learned on the DeepFakeFace has got a better performance of 50.7\% on the unseen test set than the CoaT model despite having significantly less primary test accuracy, and the Meta model has shown a 10.81\% improvement to the best performing other model on the unseen test set.
\begin{table*}[ht]
    \centering
    \caption{Performance Metrics Across Datasets}
    \begin{tabular}{@{}lccc|ccc@{}}
        \toprule
        \textbf{Dataset}  & \multicolumn{3}{c|}{\textbf{DGM}}                 & \multicolumn{3}{c}{\textbf{iFakeFaceDB}}       \\ \midrule
              \textbf{Model        $\rvert$}  & Acc    & AUC    & F1      & Acc    & AUC    & F1    \\ \midrule
        Meta model  & 0.6452 & 0.6879 & 0.6674 & 0.6806 & 0.6726 & 0.6497 \\
        ViT         & 0.4654 & 0.4971 & 0.5633 & 0.5097 & 0.4465 & 0.5318 \\
        XCiT        & 0.4571 & 0.4526 & 0.4985 & 0.4734 & 0.5135 & 0.4906 \\
        DeiT        & 0.5139 & 0.5291 & 0.5987 & 0.4895 & 0.5386 & 0.5453 \\
        Eva2        & 0.4483 & 0.4335 & 0.5142 & 0.4618 & 0.4822 & 0.4782 \\
        CoaT        & 0.4916 & 0.5437 & 0.6436 & 0.5277 & 0.5897 & 0.5901 \\ 
        \bottomrule
    \end{tabular}
\end{table*}

\begin{figure*}[ht]
  \centering
  \includegraphics[width=\linewidth]{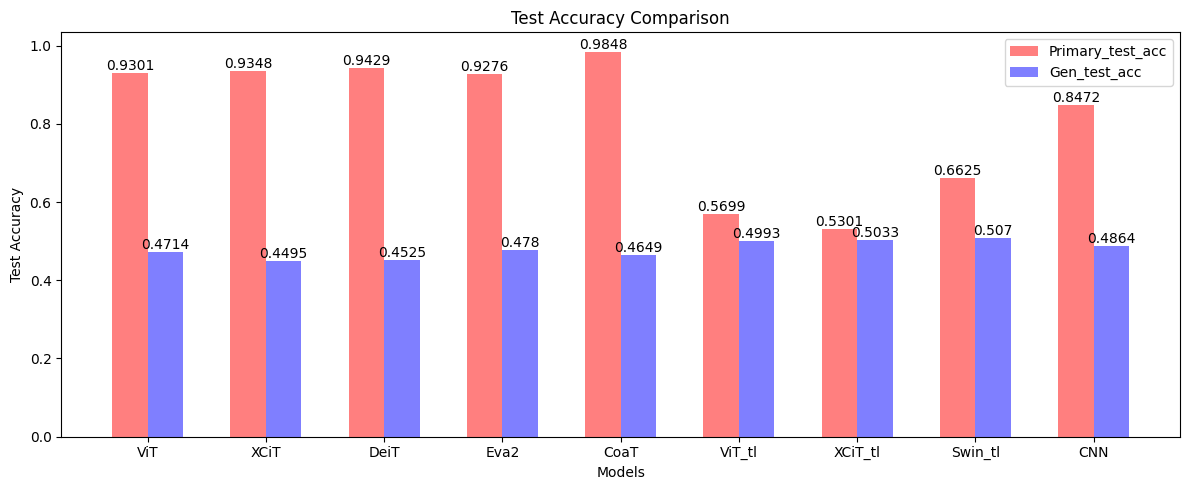}
  \caption{Comparison of test accuracy for different models on the primary test set (DeepFakeface), and on an unseen test set (OpenForensics-based)}
  \label{fig:img_11}
\end{figure*}

\item To test the meta model and the other five models on the test sets of two different datasets namely DGM, and iFakeFaceDB separately. In this case, the other five models have never seen the DGM and iFakeFaceDB datasets, nor their test sets. These other five models have been exclusively trained on DeepFakeFace dataset. The Meta model has also never seen the test sets of any of these two datasets, but has been trained on their train sets, through few-shot meta-training, where only a few train samples from one or more classes of these datasets were given for training for a task, based on N-way K-shot selection for that particular task if it includes the corresponding classes in that task. Similar to scenario-1, the comparison was made using the three metrics of accuracy, AUC, and F1 scores, and the results are presented in Table-VII. It is observed that the Meta model remained consistent in both the cases, and has a little improvement of about 3\% and 7\% respectively from the first scenario. The performance of the other models remained similar to the first scenario, with some models having a marginal improvement at cases. However, the Meta model has outperformed all the other models in both the cases.
\end{itemize}
Apparently, the Meta model trained through the proposed algorithm has shown significant improvement in the generalization capabilities across diverse datasets when compared to other similar models. The model was few-shot meta-trained in a restricted infrastructure on a system having 16 GB of RAM with 8 GB of NVIDIA GeForce RTX 4060 GDDR6 Graphic card with an Intel Core i7 $13^{th}$ gen processor. Despite few-shot training in relatively less epochs, the results show that the proposed approach has great potential in enhancing the generalization capabilities of the Meta model which could be realized if it is further trained to the full extent in the described setup with better infrastructure. \cite{10112518}

\section{\textbf{Conclusion and future scope of work}}
This work deals with the crucial challenges of model generalization, adversarial robustness, and adaptation to dynamic data drift. The problem is setup in the domain of deepfake detection. However, the considered three precise challenges go beyond deepfake detection and hold utmost importance for any classifier in any other domain or setup. 
A holistic framework with two major modules of adversarial meta-training, and hierarchical multi-agent workflow with RAG and custom image synthesis modules is proposed, designed, implemented, and experimented in this work. The proposed meta-algorithm identifies the weaknesses and strengths of the model by ranking the support set samples using the proposed metric $M_{adaptive}$ which ranks the samples in the order of Wrong predictions with large margin \> Wrong predictions with small margin \> Right predictions with small margin \> Right predictions with large margin. This helps to identify those sets of samples where the model is struggling to make a decision, and where the model is confident enough. Corresponding ensemble augmented and adversarially perturbed samples are generated and added to the support set of each of the respective tasks, in every inner step, there by making the model fix its weaknesses thus, enhancing its detection capabilities and generalization, and enhance its strengths and confidence making them adversarially robust. This way, the proposed meta-learning algorithm enhances the generalization and robustness of the model, which is demonstrated in the experimental test scenarios where the meta model trained has shown a consistent performance across different datasets. Additionally, the proposed multi-agent workflow with RAG and custom sample synthesis modules helps in collecting real-time information from diverse sources including web and news outlets, research papers and technical articles, web-based search, apart from the locally curated knowledge base. This information is processed by a hierarchy of agents, in doing various relevant tasks as described in this paper for attack pattern generation, and few-shot prompt generation, which will be subsequently used by the sample synthesis module to generate the corresponding few-shot samples pertaining to the attack pattern given or generated. The constant addition of these generated samples to the support set of tasks, makes the model realize the emerging and unforeseen trends, making it quickly adapt to newer attacks and scenarios. The paper elaborates on the working of each of the above-described modules along with the discussing their outputs and results.\\
As part of the future scope of work, the following could be some directions to begin with to sophisticate the proposed solution:
Multi-modal RAG system with Semantic chunking can be integrated into the current RAG design to get information from multi-modal inputs and more contextually relevant chunks can therefore, be made. The agents could be sophisticated by adding further agents such as for content moderation to eliminate any concerning information from processing. Besides, the agents can be better adapted to their very precise tasks assigned using instruction tuning by curating corresponding instruction datasets. The performance of the model can be further improved if a bigger architecture or variant is taken instead of the lite versions. Additionally, training them for more epochs, also, if possible, on the complete meta-dataset instead of few-shot training can all lead to improved performance of the meta-model. The current implementation primarily focusses on the image deepfakes, but its design holds true for other modalities as well. As part of the future work, it can be extended to other data modalities, making it a multi-modal solution increasing its spectrum of usecases and applications.

\bibliographystyle{ieeetr}
\bibliography{references}

\begin{thebibliography}{10}

\bibitem{article_tl}
S.~Suratkar and F.~Kazi, ``Deep fake video detection using transfer learning approach,'' {\em Arabian journal for science and engineering}, vol.~48, pp.~1--11, 10 2022.

\bibitem{9523015}
M.~Kim, S.~Tariq, and S.~S. Woo, ``Fretal: Generalizing deepfake detection using knowledge distillation and representation learning,'' in {\em 2021 IEEE/CVF Conference on Computer Vision and Pattern Recognition Workshops (CVPRW)}, pp.~1001--1012, 2021.

\bibitem{10444215}
T.~Kim, J.~Choi, H.~Cho, H.~Lim, and J.~Choi, ``Domain generalization for face forgery detection by style transfer,'' in {\em 2024 IEEE International Conference on Consumer Electronics (ICCE)}, pp.~1--5, 2024.

\bibitem{song2023robustnessgeneralizabilitydeepfakedetection}
H.~Song, S.~Huang, Y.~Dong, and W.-W. Tu, ``Robustness and generalizability of deepfake detection: A study with diffusion models,'' 2023.

\bibitem{li2024generalizabledeepfakeimagedetectors}
B.~Li, J.~Sun, C.~M. Poskitt, and X.~Wang, ``How generalizable are deepfake image detectors? an empirical study,'' 2024.

\bibitem{Passos_2024}
L.~A. Passos, D.~Jodas, K.~A.~P. Costa, L.~A. Souza~Júnior, D.~Rodrigues, J.~Del~Ser, D.~Camacho, and J.~P. Papa, ``A review of deep learning‐based approaches for deepfake content detection,'' {\em Expert Systems}, vol.~41, Feb. 2024.

\bibitem{Kamat_2023_ICCV}
S.~Kamat, S.~Agarwal, T.~Darrell, and A.~Rohrbach, ``Revisiting generalizability in deepfake detection: Improving metrics and stabilizing transfer,'' in {\em Proceedings of the IEEE/CVF International Conference on Computer Vision (ICCV) Workshops}, pp.~426--435, October 2023.

\bibitem{10495703}
M.~Ivanovska and V.~Struc, ``On the vulnerability of deepfake detectors to attacks generated by denoising diffusion models,'' in {\em 2024 IEEE/CVF Winter Conference on Applications of Computer Vision Workshops (WACVW)}, (Los Alamitos, CA, USA), pp.~1051--1060, IEEE Computer Society, jan 2024.

\bibitem{9150604}
N.~Carlini and H.~Farid, ``Evading deepfake-image detectors with white- and black-box attacks,'' in {\em 2020 IEEE/CVF Conference on Computer Vision and Pattern Recognition Workshops (CVPRW)}, pp.~2804--2813, 2020.

\bibitem{hou2023evadingdeepfakedetectorsadversarial}
Y.~Hou, Q.~Guo, Y.~Huang, X.~Xie, L.~Ma, and J.~Zhao, ``Evading deepfake detectors via adversarial statistical consistency,'' 2023.

\bibitem{10022079}
A.~Devasthale and S.~Sural, ``Adversarially robust deepfake video detection,'' in {\em 2022 IEEE Symposium Series on Computational Intelligence (SSCI)}, pp.~396--403, 2022.

\bibitem{hooda2023d4detectionadversarialdiffusion}
A.~Hooda, N.~Mangaokar, R.~Feng, K.~Fawaz, S.~Jha, and A.~Prakash, ``D4: Detection of adversarial diffusion deepfakes using disjoint ensembles,'' 2023.

\bibitem{10.1007/978-3-031-53311-2_37}
S.~Khan, J.-C. Chen, W.-H. Liao, and C.-S. Chen, ``Adversarially robust deepfake detection via adversarial feature similarity learning,'' in {\em MultiMedia Modeling: 30th International Conference, MMM 2024, Amsterdam, The Netherlands, January 29 – February 2, 2024, Proceedings, Part III}, (Berlin, Heidelberg), p.~503–516, Springer-Verlag, 2024.

\bibitem{9522903}
P.~Neekhara, B.~Dolhansky, J.~Bitton, and C.~C. Ferrer, ``Adversarial threats to deepfake detection: A practical perspective,'' in {\em 2021 IEEE/CVF Conference on Computer Vision and Pattern Recognition Workshops (CVPRW)}, pp.~923--932, 2021.

\bibitem{galdi20242dmalafideadversarialattacksface}
C.~Galdi, M.~Panariello, M.~Todisco, and N.~Evans, ``2d-malafide: Adversarial attacks against face deepfake detection systems,'' 2024.

\bibitem{9956543}
N.~Lim, M.~Y. Kuan, M.~Pu, M.~Lim, and C.~Y. Chong, ``Metamorphic testing-based adversarial attack to fool deepfake detectors,'' in {\em 2022 26th International Conference on Pattern Recognition (ICPR)}, (Los Alamitos, CA, USA), pp.~2503--2509, IEEE Computer Society, aug 2022.

\bibitem{9956520}
J.~Kim, T.~Kim, J.~Kim, and S.~S. Woo, ``Evading deepfake detectors via high quality face pre-processing methods,'' in {\em 2022 26th International Conference on Pattern Recognition (ICPR)}, (Los Alamitos, CA, USA), pp.~1937--1944, IEEE Computer Society, aug 2022.

\bibitem{10035845}
C.~Liu, H.~Chen, T.~Zhu, J.~Zhang, and W.~Zhou, ``Making deepfakes more spurious: Evading deep face forgery detection via trace removal attack,'' {\em IEEE Transactions on Dependable and Secure Computing}, vol.~20, pp.~5182--5196, nov 2023.

\bibitem{TASSONE2024104143}
F.~Tassone, L.~Maiano, and I.~Amerini, ``Continuous fake media detection: Adapting deepfake detectors to new generative techniques,'' {\em Computer Vision and Image Understanding}, vol.~249, p.~104143, 2024.

\bibitem{nichol2018firstordermetalearningalgorithms}
A.~Nichol, J.~Achiam, and J.~Schulman, ``On first-order meta-learning algorithms,'' 2018.

\bibitem{Nguyen_2022}
T.~T. Nguyen, Q.~V.~H. Nguyen, D.~T. Nguyen, D.~T. Nguyen, T.~Huynh-The, S.~Nahavandi, T.~T. Nguyen, Q.-V. Pham, and C.~M. Nguyen, ``Deep learning for deepfakes creation and detection: A survey,'' {\em Computer Vision and Image Understanding}, vol.~223, p.~103525, Oct. 2022.

\bibitem{Mirsky_2021}
Y.~Mirsky and W.~Lee, ``The creation and detection of deepfakes: A survey,'' {\em ACM Computing Surveys}, vol.~54, p.~1–41, Jan. 2021.

\bibitem{jimaging9010018}
Z.~Akhtar, ``Deepfakes generation and detection: A short survey,'' {\em Journal of Imaging}, vol.~9, no.~1, 2023.

\bibitem{mittal2022facedetectionmodelsbiased}
S.~Mittal, K.~Thakral, P.~Majumdar, M.~Vatsa, and R.~Singh, ``Are face detection models biased?,'' 2022.

\bibitem{10204714}
K.~Narayan, H.~Agarwal, K.~Thakral, S.~Mittal, M.~Vatsa, and R.~Singh, ``Df-platter: Multi-face heterogeneous deepfake dataset,'' in {\em 2023 IEEE/CVF Conference on Computer Vision and Pattern Recognition (CVPR)}, pp.~9739--9748, 2023.

\bibitem{narayan2022deephydeepfakephylogeny}
K.~Narayan, H.~Agarwal, K.~Thakral, S.~Mittal, M.~Vatsa, and R.~Singh, ``Deephy: On deepfake phylogeny,'' 2022.

\bibitem{10440475}
R.~Shao, T.~Wu, J.~Wu, L.~Nie, and Z.~Liu, ``Detecting and grounding multi-modal media manipulation and beyond,'' {\em IEEE Transactions on Pattern Analysis and Machine Intelligence}, vol.~46, no.~8, pp.~5556--5574, 2024.

\bibitem{wang2023deterdetectingeditedregions}
S.~Wang, Y.~Zhu, R.~Wang, A.~Dharmasiri, O.~Russakovsky, and Y.~Wu, ``Deter: Detecting edited regions for deterring generative manipulations,'' 2023.

\bibitem{Wang2024LinguisticPO}
Y.~Wang, Z.~Huang, Z.~Ma, and X.~Hong, ``Linguistic profiling of deepfakes: An open database for next-generation deepfake detection,'' {\em ArXiv}, vol.~abs/2401.02335, 2024.

\bibitem{9897972}
S.~Jia, X.~Li, and S.~Lyu, ``Model attribution of face-swap deepfake videos,'' in {\em 2022 IEEE International Conference on Image Processing (ICIP)}, pp.~2356--2360, 2022.

\bibitem{10208982}
S.~Jia, M.~Huang, Z.~Zhou, Y.~Ju, J.~Cai, and S.~Lyu, ``Autosplice: A text-prompt manipulated image dataset for media forensics,'' in {\em 2023 IEEE/CVF Conference on Computer Vision and Pattern Recognition Workshops (CVPRW)}, pp.~893--903, 2023.

\bibitem{Korshunov2018DeepFakesAN}
P.~Korshunov and S.~Marcel, ``Deepfakes: a new threat to face recognition? assessment and detection,'' {\em ArXiv}, vol.~abs/1812.08685, 2018.

\bibitem{9156378}
H.~Dang, F.~Liu, J.~Stehouwer, X.~Liu, and A.~K. Jain, ``On the detection of digital face manipulation,'' in {\em 2020 IEEE/CVF Conference on Computer Vision and Pattern Recognition (CVPR)}, (Los Alamitos, CA, USA), pp.~5780--5789, IEEE Computer Society, jun 2020.

\bibitem{9133490}
J.~C. Neves, R.~Tolosana, R.~Vera-Rodriguez, V.~Lopes, H.~Proença, and J.~Fierrez, ``Ganprintr: Improved fakes and evaluation of the state of the art in face manipulation detection,'' {\em IEEE Journal of Selected Topics in Signal Processing}, vol.~14, no.~5, pp.~1038--1048, 2020.

\bibitem{Li2019CelebDFAL}
Y.~Li, X.~Yang, P.~Sun, H.~Qi, and S.~Lyu, ``Celeb-df: A large-scale challenging dataset for deepfake forensics,'' {\em 2020 IEEE/CVF Conference on Computer Vision and Pattern Recognition (CVPR)}, pp.~3204--3213, 2019.

\bibitem{9010912}
A.~Rossler, D.~Cozzolino, L.~Verdoliva, C.~Riess, J.~Thies, and M.~Niessner, ``Faceforensics++: Learning to detect manipulated facial images,'' in {\em 2019 IEEE/CVF International Conference on Computer Vision (ICCV)}, (Los Alamitos, CA, USA), pp.~1--11, IEEE Computer Society, nov 2019.

\bibitem{9578592}
T.~Zhou, W.~Wang, Z.~Liang, and J.~Shen, ``Face forensics in the wild,'' in {\em 2021 IEEE/CVF Conference on Computer Vision and Pattern Recognition (CVPR)}, pp.~5774--5784, 2021.

\bibitem{9484387}
B.~Peng, H.~Fan, W.~Wang, J.~Dong, Y.~Li, S.~Lyu, Q.~Li, Z.~Sun, H.~Chen, B.~Chen, Y.~Hu, S.~Luo, J.~Huang, Y.~Yao, B.~Liu, H.~Ling, G.~Zhang, Z.~Xu, C.~Miao, C.~Lu, S.~He, X.~Wu, and W.~Zhuang, ``Dfgc 2021: A deepfake game competition,'' in {\em 2021 IEEE International Joint Conference on Biometrics (IJCB)}, pp.~1--8, 2021.

\bibitem{Zi2020WildDeepfakeAC}
B.~Zi, M.~Chang, J.~Chen, X.~Ma, and Y.-G. Jiang, ``Wilddeepfake: A challenging real-world dataset for deepfake detection,'' {\em Proceedings of the 28th ACM International Conference on Multimedia}, 2020.

\bibitem{8683164}
X.~Yang, Y.~Li, and S.~Lyu, ``Exposing deep fakes using inconsistent head poses,'' in {\em ICASSP 2019 - 2019 IEEE International Conference on Acoustics, Speech and Signal Processing (ICASSP)}, pp.~8261--8265, 2019.

\bibitem{9931802}
A.~KoÇak and M.~Alkan, ``Deepfake generation, detection and datasets: a rapid-review,'' in {\em 2022 15th International Conference on Information Security and Cryptography (ISCTURKEY)}, pp.~86--91, 2022.

\bibitem{10469620}
R.~Purohit, Y.~Sane, D.~Vaishampayan, S.~Vedantam, and M.~Singh, ``Ai vs. human vision: A comparative analysis for distinguishing ai-generated and natural images,'' in {\em 2024 Fourth International Conference on Advances in Electrical, Computing, Communication and Sustainable Technologies (ICAECT)}, pp.~1--7, 2024.

\bibitem{inbook}
J.~Bobulski and M.~Kubanek, {\em Detection of Fake Facial Images and Changes in Real Facial Images}, pp.~110--122.
\newblock Computational Collective Intelligence, 08 2024.

\bibitem{10285057}
S.~Waseem, S.~A. R.~S. Abu~Bakar, B.~A. Ahmed, Z.~Omar, T.~A.~E. Eisa, and M.~E.~E. Dalam, ``Deepfake on face and expression swap: A review,'' {\em IEEE Access}, vol.~11, pp.~117865--117906, 2023.

\bibitem{9712265}
A.~Malik, M.~Kuribayashi, S.~M. Abdullahi, and A.~N. Khan, ``Deepfake detection for human face images and videos: A survey,'' {\em IEEE Access}, vol.~10, pp.~18757--18775, 2022.

\bibitem{10112518}
R.~Chauhan, R.~Popli, and I.~Kansal, ``A systematic review on fake image creation techniques,'' in {\em 2023 10th International Conference on Computing for Sustainable Global Development (INDIACom)}, pp.~779--783, 2023.

\bibitem{10448328}
M.~M. Diniz and A.~Rocha, ``Open-set deepfake detection to fight the unknown,'' in {\em ICASSP 2024 - 2024 IEEE International Conference on Acoustics, Speech and Signal Processing (ICASSP)}, pp.~13091--13095, 2024.

\bibitem{10448251}
J.~Liu, M.~Zhang, J.~Ke, and L.~Wang, ``Advshadow: Evading deepfake detection via adversarial shadow attack,'' in {\em ICASSP 2024 - 2024 IEEE International Conference on Acoustics, Speech and Signal Processing (ICASSP)}, pp.~4640--4644, 2024.

\bibitem{inproceedings}
N.~Carlini and H.~Farid, ``Evading deepfake-image detectors with white- and black-box attacks,'' in {\em IEEE/CVF Conference on Computer Vision and Pattern Recognition Workshops (CVPRW)}, pp.~2804--2813, 06 2020.

\bibitem{10646853}
S.~Abdullah, A.~Cheruvu, S.~Kanchi, T.~Chung, P.~Gao, M.~Jadliwala, and B.~Viswanath, ``An analysis of recent advances in deepfake image detection in an evolving threat landscape,'' in {\em 2024 IEEE Symposium on Security and Privacy (SP)}, (Los Alamitos, CA, USA), pp.~91--109, IEEE Computer Society, may 2024.

\bibitem{Naitali2023DeepfakeAG}
A.~Naitali, M.~Ridouani, F.~Salahdine, and N.~Kaabouch, ``Deepfake attacks: Generation, detection, datasets, challenges, and research directions,'' {\em Comput.}, vol.~12, p.~216, 2023.

\bibitem{shao2023dgm4}
R.~Shao, T.~Wu, and Z.~Liu, ``Detecting and grounding multi-modal media manipulation,'' in {\em IEEE Conference on Computer Vision and Pattern Recognition (CVPR)}, 2023.

\bibitem{1911.05351}
J.~C. Neves, R.~Tolosana, R.~Vera-Rodriguez, V.~Lopes, H.~Proença, and J.~Fierrez, ``{GANprintR: Improved Fakes and Evaluation of the State-of-the-Art in Face Manipulation Detection},'' 2019.

\bibitem{Guillaro_2023_CVPR}
F.~Guillaro, D.~Cozzolino, A.~Sud, N.~Dufour, and L.~Verdoliva, ``Trufor: Leveraging all-round clues for trustworthy image forgery detection and localization,'' in {\em Proceedings of the IEEE/CVF Conference on Computer Vision and Pattern Recognition (CVPR)}, pp.~20606--20615, June 2023.

\bibitem{DeepfakeBench_YAN_NEURIPS2023}
Z.~Yan, Y.~Zhang, X.~Yuan, S.~Lyu, and B.~Wu, ``Deepfakebench: A comprehensive benchmark of deepfake detection,'' in {\em Advances in Neural Information Processing Systems} (A.~Oh, T.~Neumann, A.~Globerson, K.~Saenko, M.~Hardt, and S.~Levine, eds.), vol.~36, pp.~4534--4565, Curran Associates, Inc., 2023.

\bibitem{ltnghia-ICCV2021}
T.-N. Le, H.~H. Nguyen, J.~Yamagishi, and I.~Echizen, ``Openforensics: Large-scale challenging dataset for multi-face forgery detection and segmentation in-the-wild,'' in {\em International Conference on Computer Vision}, 2021.

\end{thebibliography}

\end{document}